\documentclass{article}
\usepackage{ijcai19}

\usepackage{times}
\usepackage{soul}
\usepackage{url}
\usepackage[hidelinks]{hyperref}
\usepackage[utf8]{inputenc}
\usepackage[small]{caption}
\usepackage{graphicx}
\usepackage{amsmath}
\usepackage{booktabs}
\usepackage{algorithm}
\usepackage{algorithmic}
\usepackage{amssymb}
\usepackage{diagbox}
\usepackage{multirow}
\usepackage{color}

\usepackage{amsmath,amsfonts,bm}

\def\eqref#1{equation~\ref{#1}}

\def\1{\bm{1}}

\def\vp{{\bm{p}}}
\def\vq{{\bm{q}}}

\def\vx{{\bm{x}}}

\def\mP{{\bm{P}}}

\def\mW{{\bm{W}}}

\DeclareMathAlphabet{\mathsfit}{\encodingdefault}{\sfdefault}{m}{sl}
\SetMathAlphabet{\mathsfit}{bold}{\encodingdefault}{\sfdefault}{bx}{n}

\def\gG{{\mathcal{G}}}

\def\gP{{\mathcal{P}}}

\def\gX{{\mathcal{X}}}
\def\gY{{\mathcal{Y}}}

\title{Prototype Propagation Networks (PPN) for \\Weakly-supervised Few-shot Learning on Category Graph}

\newcommand\blfootnote[1]{%
  \begingroup
  \renewcommand\thefootnote{}\footnote{#1}%
  \addtocounter{footnote}{-1}%
  \endgroup
}

\author{Lu Liu$^{1}$,
\blfootnote{This work has been accepted to IJCAI 2019. Code at: \url{https://github.com/liulu112601/PPN}.}
~Tianyi Zhou,$^{2}$~Guodong Long,$^{1}$~Jing Jiang,$^{1}$~Lina Yao,$^{3}$~Chengqi Zhang$^{1}$\\
\normalfont \small $^{1}$Center for Artificial Intelligence, FEIT, University of Technology Sydney\\$^{2}$Paul G. Allen School of Computer Science \& Engineering, University of Washington\\$^{3}$School of Computer Science and Engineering, University of New South Wales\\
{\tt\small lu.liu-10@student.uts.edu.au},~{\tt\small tianyizh@uw.edu},~{\tt\small guodong.long@uts.edu.au},\\~{\tt\small jing.jiang@uts.edu.au},~{\tt\small lina.yao@unsw.edu.au},~{\tt\small chengqi.zhang@uts.edu.au}
}

\begin{document}

\maketitle

\begin{abstract}
A variety of machine learning applications expect to achieve rapid learning from a limited number of labeled data. 
However, the success of most current models is the result of heavy training on big data. 
Meta-learning addresses this problem by extracting common knowledge across different tasks that can be quickly adapted to new tasks. 
However, they do not fully explore weakly-supervised information, which is usually free or cheap to collect. 
In this paper, we show that weakly-labeled data can significantly improve the performance of meta-learning on few-shot classification. 
We propose \textit{prototype propagation network (PPN)} trained on few-shot tasks together with data annotated by coarse-label.
Given a category graph of the targeted fine-classes and some weakly-labeled coarse-classes, PPN learns an attention mechanism which propagates the prototype of one class to another on the graph, so that the K-nearest neighbor (KNN) classifier defined on the propagated prototypes results in high accuracy across different few-shot tasks.
The training tasks are generated by subgraph sampling, and the training objective is obtained by accumulating the level-wise classification loss on the subgraph. 
The resulting graph of prototypes can be continually re-used and updated for new tasks and classes. 
We also introduce two practical test/inference settings which differ according to whether the test task can leverage any weakly-supervised information as in training.
On two benchmarks, PPN significantly outperforms most recent few-shot learning methods in different settings, even when they are also allowed to train on weakly-labeled data.
\end{abstract}

\section{Introduction}

\begin{figure}[t!]
\begin{center}
\includegraphics[width=\linewidth]{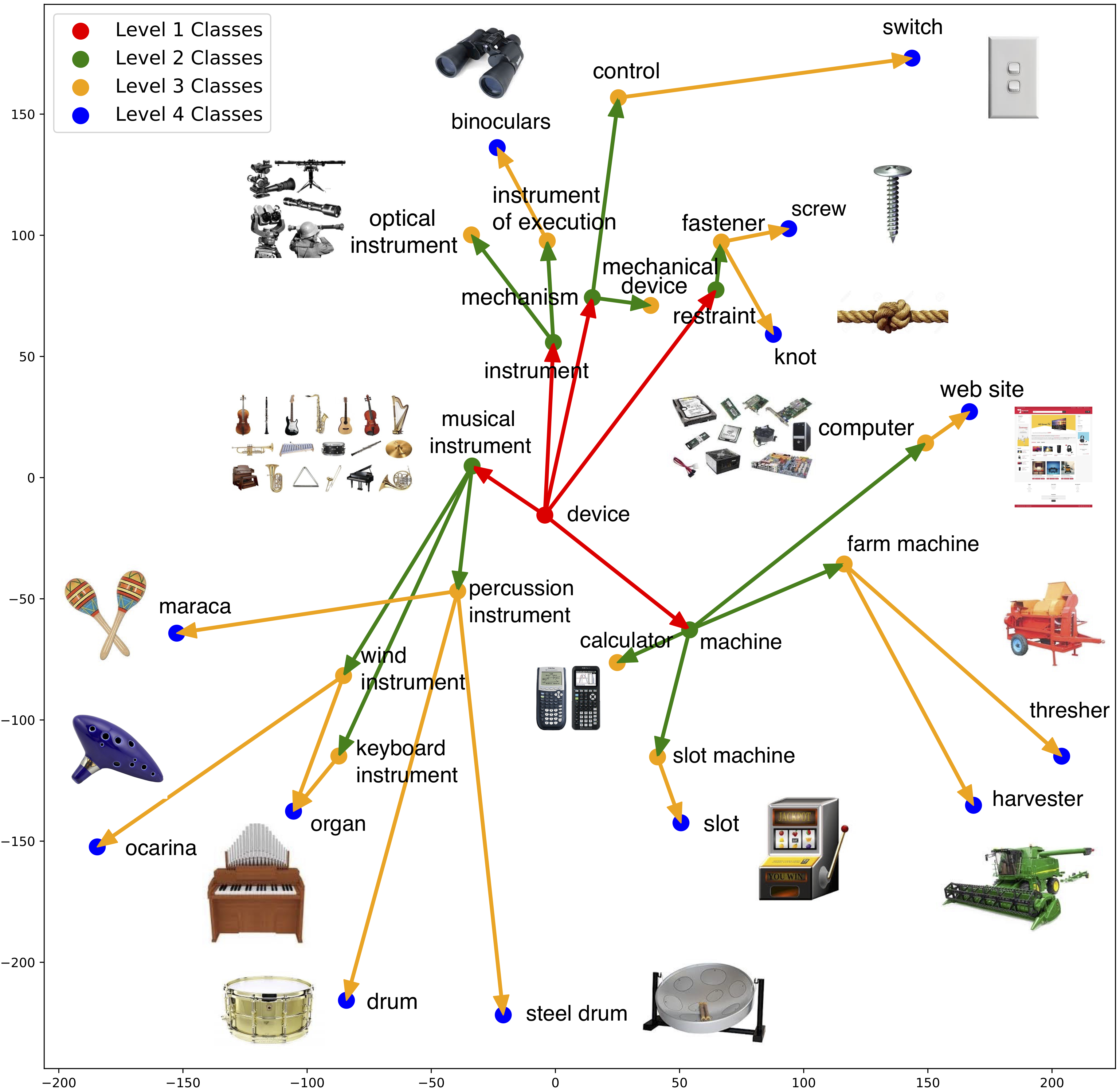}
\end{center}
\caption{
Prototypes learned by PPN and transformed to a 2D space by t-SNE~\protect\cite{maaten2008visualizing}. Each edge connects a child class to a parent. The prototypes spread out as classes become finer, preserve the graph structure, and reflect the semantic similarity.}
\label{fig:t-sne}
\end{figure}

Machine learning (ML) has achieved breakthrough success in a great number of application fields during the past 10 years, due to more expressive model structures, the availability of massive training data, and fast upgrading of computational hardware/infrastructure. 
Nowadays, with the support of expensive hardware, we can train super-powerful deep neural networks containing thousands of layers on millions or even trillions of data within an acceptable time. 
However, as AI becomes democratized for personal or small business use, with concerns about data privacy, demand is rapidly growing for instant learning of highly customized models on edge/mobile devices with limited data. 
This brings new challenges since the big data and computational power that major ML techniques rely on are no longer available or affordable. 
In such cases, ML systems that can quickly adapt to new tasks and produce reliable models by only seeing few-shot training data are highly preferable.

\begin{figure*}[t]
\begin{center}
\includegraphics[width=\linewidth]{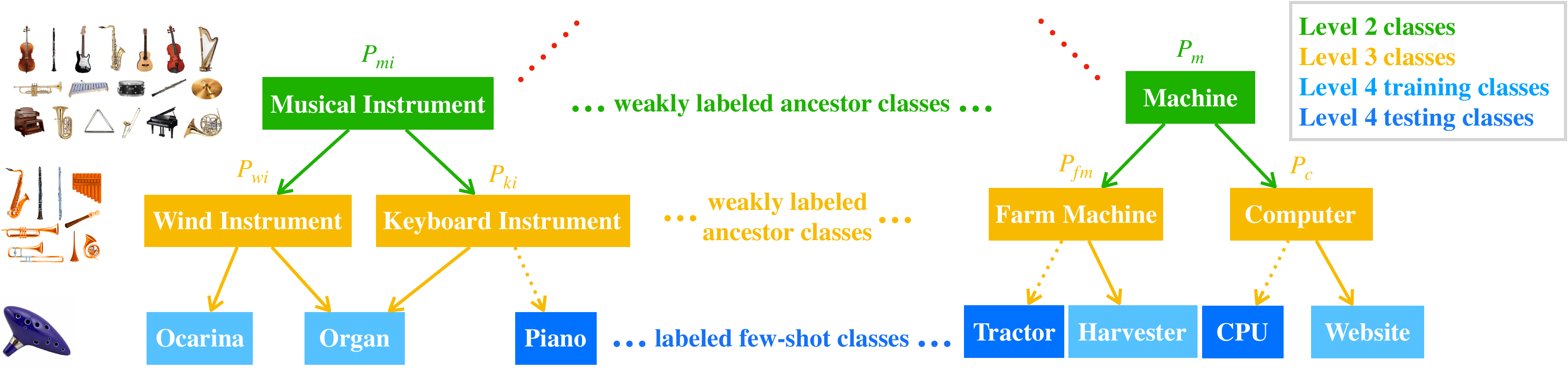}
\end{center}
\caption{Weakly-supervised few-shot learning in PPN. 
The few-shot classes (leaf nodes) in training and test tasks are non-overlapping, but they are allowed to share some parent classes.  Weakly-labeled data only has non-leaf class labels.
As shown in the left side, finer class labels are more informative but more expensive to collect, so we assume that the number of weakly-labeled data exponentially reduces as their labels become finer. 
PPN is trained on classification tasks with both fine and coarse classes.
}
\label{fig:weakly-sueprvised}
\end{figure*}

This few-shot learning problem can be addressed by a class of approaches called ``meta-learning''. Instead of independently learning each task from scratch, the goal of meta-learning is to learn the common knowledge shared across different tasks, or ``learning to learn''.
The knowledge is at learning/algorithm-level and is task-independent, and thus can be applied to new unseen tasks.
For example, it can be shared initialization weights~\cite{finn2017model}, an optimization algorithm~\cite{ravi2017optimization}, a distance/similarity metric~\cite{vinyals2016matching}, or a generator of prototypes~\cite{snell2017prototypical} that compose the support set of the K-nearest neighbor (KNN) predictor. Therefore, new tasks can benefit from the accumulated meta-knowledge extracted from previous tasks. In contrast to single-task learning, the ``training data'' in meta-learning are tasks, i.e., tasks from a certain distribution are sampled across all possible tasks. It then tries to maximize the validation accuracy of these sampled tasks. Meta-learning shares ideas with life-long/continual/progressive learning in that the meta-knowledge can be re-used and updated for future tasks. It generalizes multi-task learning~\cite{Caruana1997} since it can be applied to any new task drawn from the same distribution. 

Although recent studies of meta-learning have shown its effectiveness on few-shot learning tasks, most of them do not leverage weakly-supervised information, which is usually free or cheap to collect, and has been proved to be helpful when training data is insufficient, e.g., in weakly-supervised~\cite{zhou2017brief} and semi-supervised learning~\cite{belkin2006manifold,zhu2002learning}.
In this paper, we show that weakly-supervised information can significantly improve the performance of meta-learning on few-shot classification. In particular, we leverage weakly-labeled data that are annotated by coarse-class labels, e.g., an image of a Tractor with a coarse label of Machine. These data are usually cheap and easy to collect from web tags or human annotators. We additionally assume that a category graph describing the relationship between fine classes and coarse classes is available, where each node represents a class and each directed edge connects a fine-class to its parent coarse-class. An example of a  category graph is given in Figure~\ref{fig:weakly-sueprvised}.
It is not necessary for the graph to cover all the possible relationships and graph containing partial relationship is usually easy to obtain,
e.g., the ImageNet category tree built based on the synsets from WordNet.

We propose a meta-learning model called prototype propagation network (PPN) to explore the above weakly-supervised information for few-shot classification tasks. PPN produces a prototype per class by propagating the prototype of each class to its child classes on the category graph, where an attention mechanism generates the edge weights used for propagation. The learning goal of PPN is to minimize the validation errors of a KNN classifier built on the propagated prototypes for few-shot classification tasks. The prototype propagation on the graph enables the classification error of data belonging to any class to be  back-propagated to the prototype of another class, if there exists a path between the two classes.
The classification error on weakly-labeled data can thus be back-propagated to improve the prototypes of other classes, which later contribute to the prototype of few-shot classes via prototype propagation. Therefore, the weakly-labeled data and coarse classes can directly contribute to the update of prototypes of few-shot fine classes, offsetting the lack of training data. The resulting graph of prototypes can be repeatedly used, updated and augmented on new tasks/classes as an episodic memory. Interestingly, this interplay between the prototype graph and the quick adaptation to new few-shot tasks (via KNN) is analogous to the complementary learning system that reconciles episodic memory with statistical learning inside the human hippocampus~\cite{CLS}, which is believed to be critical for rapid learning.

Similar to other meta-learning methods, PPN learns from the training processes of different few-shot tasks, each defined on a subset of classes sampled from the category graph. To fully explore the weakly-labeled data,
we develop a level-wise method to train tasks, generated by subgraph sampling, for both coarse and fine classes on the graph.
In addition, we introduce two testing/inference settings that are common in different practical application scenarios: one (PPN+) is allowed to use weakly-labeled data in testing tasks and is given the edges connecting test classes to the category graph, while the other (PPN) cannot access any extra information except for the few-shot training data of the new tasks. In experiments, we extracted two benchmark datasets from ILSVRC-12 (ImageNet)~\cite{imagenet}, specifically for weakly-supervised few-shot learning. In different test settings, our method consistently and significantly outperforms the three most recently proposed few-shot learning models and their variants, which also trained on weakly-labeled data. The prototypes learned by PPN is visualized in Figure~\ref{fig:t-sne}.


\section{Related Works}

A great number of meta-learning approaches have been proposed to address the few-shot learning problem. There are usually two main ideas behind these works: 1) learning a component of a KNN predictor applied to all the tasks, e.g., the support set~\cite{snell2017prototypical}, the distance metric~\cite{vinyals2016matching}, or both~\cite{mishra2018snail}; 2) learning a component of an optimization algorithm used to train different tasks, e.g., an initialization point~\cite{finn2017model}. Another straightforward approach is to generate more training data for few-shot tasks by a data augmentation technique or generative model~\cite{lake2015human,wong2015one}. Our method follows the idea of learning a small support set (prototypes) for KNN, and differs in that we leverage the weakly-labeled data by relating prototypes of different classes. 

Auxiliary information: unlabeled data~\cite{ren2018meta} and inter/intra-task relationship~\cite{nichol2018reptile,liu2018transductive,amortized} have recently been used to improve the performance of few-shot learning.
Co-training on auxiliary tasks~\cite{oreshkin2018tadam} has also been applied to improve the learning of the similarity metric. In contrast, to the best of our knowledge, we are the first to utilize the weakly-labeled data as the auxiliary information to bring significant improvement.

The prototype propagation in our method inherits ideas from random walk, message passing, belief propagation, and label propagation~\cite{wu2019comprehensive,zhou2016scaling,dong2019search}. A similar idea has also been used in more recent graph neural networks (GNN) such as graph attention networks (GAT)~\cite{velivckovic2018graph}. GNN are mainly designed for tasks on graph-structured data such as node classification~\cite{hamilton2017inductive}, graph embedding~\cite{yu2018learning}, graph clustering~\cite{daegc} and graph generation~\cite{dai2018syntax}. Although our method uses an attention mechanism similar to GAT for propagation, we have a different training scheme (Algorithm~\ref{alg:level-train}) that only requires one-step propagation on a specific directed acyclic graph (DAG).
GNN has been applied to meta-learning in~\cite{gnn2018few}, but the associated graph structure is defined on samples (images) instead of classes/prototypes, and is handcrafted with fully connected edges.


\begin{figure*}[!ht]
\begin{center}
\includegraphics[width=\linewidth]{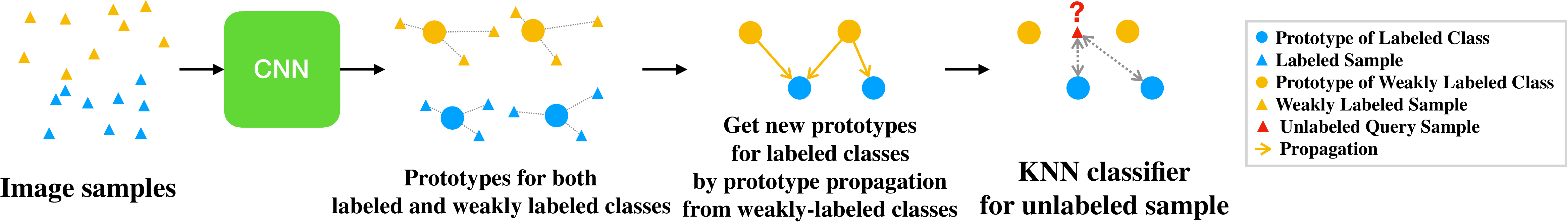}
\end{center}
\caption{An overview for few-shot classifications using PPN. Prototypes are initialized by averaging sample representations (provided by a CNN) per class. They are then updated by weighted averaging of the prototypes of their parent classes (weakly-labeled classes).
The above process generates a learner model (i.e., a KNN): the class whose prototype is the closest to a query sample is the output prediction.}
\label{fig:computation-graph}
\end{figure*}
 
\section{Prototype Propagation Networks (PPN)}

\begin{table}[t]
\centering
\caption{Notations used in this paper.}
\begin{tabular}{ll}
\toprule
Notation & Definition \\ 
\midrule
$\mathcal X$ & Data for targeted fine-classes \\
$\mathcal Y$ & Labels for targeted fine-classes \\
$\mathcal X^{w}$ & Data for weakly-labeled classes \\
$\mathcal Y^{w}$ & Labels for weakly labeled classes \\
$\mathcal G$ & Category graph (a directed acyclic graph) \\
$E$ & Edges of $\mathcal G$\\
$\mathcal G_{i}^{j}$ & Level-$j$ of $\mathcal G_{i}$ (a subgraph of $
\mathcal G$) \\
$D^{\mathcal G^j_i}$ & Distribution of data from classes on subgraph $\mathcal G_{i}^{j}$ \\
$\mathcal Y^{\mathcal G_i}$ & Set of classes on subgraph $\mathcal G_{i}$ \\
$\mathcal Y^{\mathcal G^j_i}$ & Set of classes from level-$j$ of subgraph $\mathcal G_{i}$ \\
$T$ & a few-shot learning task \\
$\mathcal T$ & Distribution that task $T$ is drawn from \\
$D^{T}$ & Distribution of data from classes in task $T$ \\
$w$ & a meta-learner producing learner models\\
$\Theta$ & Parameters of a meta learner \\
$\mP_{y}$ & Final prototype for class $y$\\
$\mP^{0}_{y}$ & Initialized Prototype for class $y$\\
$\mP^{+}_{y}$ & Propagated Prototype for class $y$\\
\bottomrule
\end{tabular}
\end{table}


\subsection{Weakly-Supervised Few-Shot Learning}\label{sec:problem}

In weakly-supervised few-shot learning, we learn from two types of data: the few-shot data $\mathcal X$ annotated by the target fine-class labels and the weakly-labeled data $\mathcal X^w$ annotated by coarse-class labels. 
Each $x\in\mathcal X$ is associated with a fine-class label $y \in\mathcal Y$, while each $x\in\mathcal X^w$ is associated with a coarse-class label $y^w \in\mathcal Y^w$. We assume that there exists a category graph describing the relationship between fine classes and coarse classes. This is a directed acyclic graph (DAG) $\mathcal G=(\mathcal Y\cup \mathcal Y^w, E)$, where each node $y\in \mathcal Y\cup \mathcal Y^w$ denotes a class, and each edge (or arc) $z\rightarrow y\in E$ connects a parent class $z$ to one of its child classes $y$. An example of the category graph is given in Figure~\ref{fig:weakly-sueprvised}: the few-shot classes $\mathcal Y$ are the leaves of the graph, while the weakly-labeled classes $\mathcal Y^w$ are the parents \footnote{\textbf{Parents}: directly connected coarse classes.} or ancestors \footnote{\textbf{Ancestors}: coarse classes not connected but linked via paths.} of the few-shot classes; for example, the parent class of ``Tractor'' is ``Farm Machine'', while its ancestors include ``Farm Machine'' and ``Machine''. A child class can belong to multiple parent classes, e.g., the class ``Organ'' has two parent classes, ``Wind Instrument'' and ``Keyboard Instrument''.

We follow the setting of few-shot learning, which draws training tasks $T\sim\mathcal T$ from a task distribution $\mathcal T$ and assumes that the test tasks are also drawn from the same distribution. For few-shot classification, each task $T$ is defined by a subset of classes, e.g., an $N$-way $k$-shot task refers to classification over $N$ classes and each class only has $k$ training samples. 
It is necessary to sample the training classes and test classes from two disjoint sets of classes to avoid overlapping.
In our problem, as shown by Figure~\ref{fig:weakly-sueprvised}, the few-shot classes used for training and test (colored light blue and deep blue respectively) are also non-overlapping, but we allow them to share some parents on the graph. We also allow training tasks to cover any classes on the graph.
Since finer-class labels can provide more information about the targeted few-shot classes but are more expensive to obtain, we assume that the amount of weakly-labeled data is reduced exponentially when the class becomes finer. The training aims to solve the following risk minimization (or likelihood maximization) of ``learning to learn'':
\begin{equation}\label{equ:opt-obj}
\min_\Theta\mathbb E_{T\sim\mathcal T}\left[\mathbb E_{(x,y)\sim\mathcal D^T}  -\log\Pr(y|x; w(T;\Theta))\right],
\end{equation}
\noindent where each task $T$ is defined by a subset of classes $\mathcal Y^T\subseteq \mathcal Y\cup \mathcal Y^w$, $\mathcal D^T$ is the distribution of data-label pair $(x,y)$ with $y\in \mathcal Y^T$, $\Pr(y|x; w(T;\Theta))$ is the likelihood of $(x,y)$ produced by model $w(T;\Theta)$ for task $T$, where $\Theta$ is the meta-learner parameter shared by all the tasks drawn from $\mathcal T$.

In our method, $\Pr(y|x; w(T;\Theta))$ is computed by a soft-version of KNN, where $w(T;\Theta)$ is the set of nearest neighbor candidates (i.e., the support set of KNN) 
for task $T$, and $\Theta$ defines the mechanism generating the support set. 
Similar to prototypical networks \cite{snell2017prototypical}, the support set $w(T;\Theta)$ 
is composed of prototypes $\mP_{\mathcal Y^T}\triangleq\{\mP_{y}\in\mathbb R^d:y\in\mathcal Y^T\}$, each of which is associated with a class $y\in\mathcal Y^T$. Given a data point $x$, we first compute its representation $f(x)\in\mathbb R^d$, where $f(\cdot)$ is convolutional neural networks (CNN) with parameter $\Theta^{cnn}$, then $\Pr(y|x; w(T;\Theta))=\Pr(y|\vx;\mP_{\mathcal Y^T})$ is computed as:
\begin{equation}\label{equ:euc-dis}
\Pr(y|\vx;\mP_{\mathcal Y^T}) \triangleq \frac{\exp(-\|f(\vx)-\mP_{y})\|^2)}{\sum_{z\in \mathcal Y^T}\exp(-\|f(\vx)-\mP_{z})\|^2)},
\end{equation}
In the following, we will introduce prototype propagation which generates $\mP_{\mathcal Y^T}$ for any task $T\sim\mathcal T$. An overall of all procedures in our model is shown in Figure~\ref{fig:computation-graph}.

\subsection{Prototype Propagation}\label{sec:MPG}

In PPN, each training task $T$ is a level-wise classification on a sampled subgraph $\mathcal G_i\subseteq \mathcal G$, i.e., a classification task over $\mathcal Y^T=\mathcal Y^{\mathcal G^j_i}$, where $\mathcal G_i^j$ denotes the level-$j$ of subgraph $\mathcal G_i$ and $\mathcal Y^{\mathcal G^j_i}$ is the set of all the classes on $\mathcal G_i^j$. 

The prototype propagation is defined on each subgraph $\mathcal G_i$, which covers classes $\mathcal Y^{\mathcal G_i}$. Given the associated training data $\mathcal X^y$ for class $y\in \mathcal Y^{\mathcal G_i}$, the prototype of the class $y$ is initialized as the average of the representations $f(\vx)$ of the samples $\vx\in\mathcal X^y$, i.e.,
\begin{align}\label{equ:proto-graph}
\mP^0_{y}\triangleq\frac{1}{|\mathcal X^y|} \sum_{\vx\in \mathcal X^y}f(\vx).
\end{align}
For each parent class $z$ of class $y$ on $\gG_{i}$, 
we propagate $\mP^0_{z}$ to class $y$ with edge weight $a(\mP^0_{y}, \mP^0_{z})$ measuring the similarity between class $y$ and $z$, and aggregate the propagation (the messages) from all the parent classes by
\begin{equation}\label{equ:p-par-hop1}
    \mP_{y}^{+}\triangleq\sum_{z\in\mathcal P_{y}^{\mathcal G_i}} a(\mP^0_{y}, \mP^0_{z}) \times \mP^0_{z},
\end{equation}
where $\mathcal P_{y}^{\mathcal G_i}$ denotes the set of all parent classes of $y$ on the subgraph $\mathcal G_i$, and the edge weight $a(\mP^0_{y}, \mP^0_{z})$ is a learnable similarity metric defined by a dot-product attention~\cite{vaswani2017attention}, i.e.,
\begin{equation}\label{equ:att-score}
    a(\vp, \vq)\triangleq\frac{\langle g(\vp), h(\vq)\rangle}{\|g(\vp)\| \times \|h(\vq)\|},
\end{equation}
where $g(\cdot)$ and $h(\cdot)$ are learnable transformations applied to prototypes with parameters $\Theta^{att}$, e.g., linear transformations $g(\vp)=\mW_g\vp$ and $h(\vq)=\mW_h\vq$.
The prototype after propagation is a weighted average of $\mP^0_{y}$ and $\mP_{y}^{+}$ with weight $\lambda\in \left[0,1\right]$, i.e., 
\begin{align}
\label{equ:p-final}
\mP_{y}\triangleq\lambda \times \mP^0_{y} + (1-\lambda) \times \mP^{+}_{y}.
\end{align}
For each classification task $T$ on subgraph $\gG_i$, $\mP_{y}$ is used in Eq.~(\ref{equ:euc-dis}) to produce the likelihood probability. The likelihood maximization in Eq.~(\ref{equ:opt-obj}) aims to optimize the parameters $\Theta^{cnn}$ from $f(\cdot)$ and $\Theta^{att}$ from the attention mechanism across all the training tasks.

\subsection{Level-wise Training of PPN on Subgraphs}
\label{sec:training-strategy}

The goal of meta-training is to learn a parameterized propagation mechanism defined by Eq.~(\ref{equ:proto-graph})-Eq.~(\ref{equ:p-final}) on few-shot tasks. In each iteration, we randomly sample a subset of few-shot classes, which together with all their parent classes and edges on $\gG$ form a subgraph $\gG_i$. A training task $T$ is drawn from each level $\mathcal G_i^j\sim \gG_i$ as the classification over classes $\mathcal Y^T=\mathcal Y^{\mathcal G^j_i}$. The meta-learning in Eq.~(\ref{equ:opt-obj}) is approximated by
\begin{equation}
    \min_{\Theta^{cnn}, \Theta^{att}}\sum_{\mathcal G_i\sim\mathcal G}\sum_{\mathcal G^j_i\sim\mathcal G_i}\sum_{(x,y)\sim\mathcal D^{\mathcal G^j_i}}  -\log\Pr(y|x; \mP_{\mathcal Y^{\mathcal G^j_i}}),
\end{equation}
where $\mathcal D^{\mathcal G^j_i}$ is the data distribution of data-label pair $(x,y)$ from classes $\mathcal Y^{\mathcal G^j_i}$. Since the prototype propagation is defined on the whole subgraph, it generates a computational graph relating each class to all of its parent classes. Hence, during training, the classification error on each class is back-propagated to the prototypes of its parent classes, which will be updated to improve the validation accuracy of finer classes and propagated to generate the prototypes of the few-shot classes later. Therefore, the weakly-labeled data of a coarse class will contribute to the few-shot learning tasks on the leaf-node classes.



The complete level-wise training procedure of PPN is given in Algorithm~\ref{alg:level-train}, each of whose iterations comprises two main stages: prototype propagation (lines 9-11) which builds a computational graph over prototypes of the classes on a sampled subgraph $\gG_i$, and level-wise training (lines 12-16) which updates the parameters $\Theta^{cnn}$ and $\Theta^{att}$ on per-level classification task.
In the prototype propagation stage, the prototype of each class is merged with the information from the prototypes of its parent classes, and the classification error using the merged prototypes will be backpropagated to update the parent prototypes during the level-wise training stage. 
To improve the computational efficiency, we do not update $\mP^0_{y}$ for every propagation. Instead, we lazily update $\mP^0_{y}$ for all classes $y\in\gY\cup\gY^{w}$ every $m$ epochs, as shown in lines 3-7.

\begin{algorithm}[t!]
\caption{Level-wise Training of PPN}
\label{alg:level-train}
\begin{algorithmic}[1]
\REQUIRE few-shot data $\gX$ with labels from $\gY$,\\
\hspace{4mm} weakly-labeled data $\gX^{w}$ with labels from $\gY^{w}$, \\
\hspace{4mm} category graph $\gG=(\gY\cup\gY^{w}, E)$, \\
\hspace{4mm} learning rate scheduler for SGD, $\lambda$, $m$;
\STATE {\bf Initialize:} randomly initialize $\mP$, $\Theta^{cnn}$, $\Theta^{att}$, $\tau\leftarrow 0$;
\WHILE{not converge}
\IF{$\tau \bmod m= 0$}
\FOR{class $y\in \gY\cup\gY^{w}$}
\STATE  Lazily update $\mP^0_{y}$ by Eq.~(\ref{equ:proto-graph}) and save it in buffer;
\ENDFOR
\ENDIF
\STATE Sample a subgraph $\gG_{i}\sim \gG$;
\FOR{class $y\in\mathcal Y^{\gG_{i}}$}
\STATE Get $\mP^0_{y}$ from buffer, and propagation by Eq.~(\ref{equ:p-par-hop1})-(\ref{equ:p-final});
\ENDFOR
\STATE initialize loss $L\leftarrow 0$;
\FOR{level-$j$ $\mathcal G^j_i\sim\mathcal G_i$}
\STATE $L\leftarrow L+\sum
\limits_{(x,y)\sim\mathcal D^{\mathcal G^j_i}} -\log\Pr(y|x; P_{\mathcal Y^{\mathcal G^j_i}})$ by Eq.~(\ref{equ:euc-dis});
\ENDFOR
\STATE Mini-batch SGD to minimize $L$, update $\Theta^{cnn}$, $\Theta^{att}$;
\STATE $\tau\leftarrow \tau+1$;
\ENDWHILE
\end{algorithmic}
\end{algorithm}

\begin{table*}[t]
\caption{
Statistics of WS-ImageNet-Pure and WS-ImageNet-Mix for weakly-supervised few-shot learning, where \#classes and \#img denote the number of classes and images respectively. Their classes are extracted from levels 3-7 of ImageNet WordNet hierarchy. 
}
\hspace{0.5em}
\setlength{\tabcolsep}{8pt}
\begin{tabular}{lllllllllllllll}
\cline{1-11}
\multicolumn{1}{|l|}{}   & \multicolumn{5}{c|}{WS-ImageNet-Pure}
& \multicolumn{5}{c|}{WS-ImageNet-Mix}  &  &  \\ \cline{1-11}
\multicolumn{1}{|l|}{\multirow{2}{*}{level}} & \multicolumn{2}{c|}{training}                               & \multicolumn{2}{c|}{testing}                                & \multicolumn{1}{l|}{\multirow{2}{*}{\#img}}  & \multicolumn{2}{c|}{training}                               & \multicolumn{2}{c|}{testing}                                & \multicolumn{1}{l|}{\multirow{2}{*}{\#img}} &   &  \\ \cline{2-5} \cline{7-10}
\multicolumn{1}{|l|}{}                       & \multicolumn{1}{l|}{\#classes} & \multicolumn{1}{l|}{\#img} & \multicolumn{1}{l|}{\#classes} & \multicolumn{1}{l|}{\#img} & \multicolumn{1}{l|}{}                          & \multicolumn{1}{l|}{\#classes} & \multicolumn{1}{l|}{\#img} & \multicolumn{1}{l|}{\#classes} & \multicolumn{1}{l|}{\#img} & \multicolumn{1}{l|}{}                          &  &  \\ \cline{1-11}

\multicolumn{1}{|c|}{3}                      & \multicolumn{1}{l|}{7}          & \multicolumn{1}{l|}{15140}      & \multicolumn{1}{l|}{1}          & \multicolumn{1}{l|}{3680}      & \multicolumn{1}{l|}{18820}  & \multicolumn{1}{l|}{7}          & \multicolumn{1}{l|}{92785}      & \multicolumn{1}{l|}{4}          & \multicolumn{1}{l|}{42882}      & \multicolumn{1}{l|}{135667}                      &  &  \\ \cline{1-11}

\multicolumn{1}{|c|}{4}                      & \multicolumn{1}{l|}{13}          & \multicolumn{1}{l|}{10093}      & \multicolumn{1}{l|}{6}          & \multicolumn{1}{l|}{2662}      & \multicolumn{1}{l|}{12755}    & \multicolumn{1}{l|}{13}          & \multicolumn{1}{l|}{76889}      & \multicolumn{1}{l|}{7}          & \multicolumn{1}{l|}{24592}      & \multicolumn{1}{l|}{101481}                         &  &  \\ \cline{1-11}

\multicolumn{1}{|c|}{5}                      & \multicolumn{1}{l|}{25}          & \multicolumn{1}{l|}{6904}      & \multicolumn{1}{l|}{9}          & \multicolumn{1}{l|}{1834}      & \multicolumn{1}{l|}{8738}                         & \multicolumn{1}{l|}{23}          & \multicolumn{1}{l|}{50348}      & \multicolumn{1}{l|}{12}          & \multicolumn{1}{l|}{13462}      & \multicolumn{1}{l|}{63810}                         &  &  \\ \cline{1-11}

\multicolumn{1}{|c|}{6}                      & \multicolumn{1}{l|}{44}          & \multicolumn{1}{l|}{4350}      & \multicolumn{1}{l|}{18}          & \multicolumn{1}{l|}{1155}      & \multicolumn{1}{l|}{5505}                         & \multicolumn{1}{l|}{42}          & \multicolumn{1}{l|}{22276}      & \multicolumn{1}{l|}{18}          & \multicolumn{1}{l|}{5849}      & \multicolumn{1}{l|}{28125}                          &  &  \\ \cline{1-11}

\multicolumn{1}{|c|}{7}                      & \multicolumn{1}{l|}{71}          & \multicolumn{1}{l|}{2538}      & \multicolumn{1}{l|}{18}          & \multicolumn{1}{l|}{646}      & \multicolumn{1}{l|}{3184}                         & \multicolumn{1}{l|}{71}          & \multicolumn{1}{l|}{7546}      & \multicolumn{1}{l|}{18}          & \multicolumn{1}{l|}{1919}      & \multicolumn{1}{l|}{9465}                         &  &  \\ \cline{1-11}

\multicolumn{1}{|c|}{all}                      & \multicolumn{1}{l|}{160}          & \multicolumn{1}{l|}{39025}      & \multicolumn{1}{l|}{52}          & \multicolumn{1}{l|}{9977}      & \multicolumn{1}{l|}{49002}                         & \multicolumn{1}{l|}{156}          & \multicolumn{1}{l|}{249844}      & \multicolumn{1}{l|}{59}          & \multicolumn{1}{l|}{88704}      & \multicolumn{1}{l|}{338548}                          &  &  \\ \cline{1-11}
\label{tab:datasets}
\end{tabular}
\end{table*}

\subsection{Meta-Test: Apply PPN to New Tasks}

We study two test settings for weakly-supervised few-shot learning, both of which will be used in the evaluation of PPN in our experiments. They differ in whether or not the weakly-supervised information, i.e., the weakly-labeled data and the connections of new classes to the category graph, is still accessible in the test tasks.
The first setting (PPN+) is allowed to access this information within test tasks while the second setting (PPN) is unable to access the information.
In other words, for PPN+, the edges (i.e., the yellow arrows in Figure~\ref{fig:computation-graph}) between each test class and any other classes are known during test phase, while these edges are unknown and needed to be predicted in the PPN setting.
The second setting is more challenging but is preferred in a variety of applications, for example, where the test tasks happen on different devices, whereas the first setting is more appropriate for life-long/continual learning on a single machine.


In the second setting, we can still leverage the prototypes achieved in the training phase and use them for the propagation of prototypes for test classes. In particular, for an unseen test class $y$ (and its associated samples $\gX^y$), we find the $K$-nearest neighbors of $\mP^0_y$ among all the prototypes achieved during training, and treat the training classes associated with the $K$-nearest neighbor prototypes as the parents of $y$ on $\gG$. These parent-class prototypes serve to provide weakly-supervised information to the test classes via propagation on the category graph.

In both settings, for each class $y$ in a test task $T$, we apply the prototype propagation in Eq.~(\ref{equ:proto-graph})-(\ref{equ:p-final}) on a subgraph composed of $y$ and its parents $\gP^{\gG}_{y}$. This produces the final prototype $\mP_y$, which is one of the candidates for the nearest neighbors for $\mP_{\gY^T}$ in KNN classification on task $T$.
In the first setting (PPN+), the hierarchy covering both the training and test classes is known so the parent class of each test class $y$ may come from either the training classes or the weakly-labeled test classes. 
When a parent class $y'\in\gP^{\gG}_{y}$ is among the training classes, we use the buffered prototype $\mP^0_{y'}$ from training for propagation; otherwise, we use Eq.~(\ref{equ:proto-graph}) to compute $\mP^0_{y'}$ over weakly-labeled samples belonging to class $y'$ for this task. 
In the second setting (PPN), since the edges connecting test classes are unknown and are predicted by KNN as introduced in the last paragraph, we assume that all the parents of $y$ are from training classes.
We directly use the parents' buffered prototypes from training for propagation.

\begin{table*}
\centering
\setlength{\tabcolsep}{3pt}
\caption{Validation accuracy (mean$\pm$CI$\%95$) on $600$ test tasks of PPN/PPN+ and baselines on WS-ImageNet-Pure.}
\begin{tabular}{lccccccc}
\toprule
\textbf{Model}            & \textbf{Weakly-Supervised}   & \textbf{5way1shot} & \textbf{5way5shot} & \textbf{10way1shot} & \textbf{10way5shot}  \\ \hline
Prototypical Net~\cite{snell2017prototypical} & N & 33.17$\pm$1.65\% & 46.76$\pm$0.98\% & 20.48$\pm$0.99\% & 31.49$\pm$0.57\% \\
GNN~\cite{gnn2018few} & N & 30.83$\pm$0.66\% & 41.33$\pm$0.62\% & 20.33$\pm$0.60\% & 22.50$\pm$0.62\% \\
Closer Look~\cite{closerlook}  & N & 32.27$\pm$1.58\% & 46.02$\pm$0.74\% & 22.78$\pm$0.94\% & 28.04$\pm$0.36\% \\
\midrule
Prototypical Network*  & Y  & 32.13$\pm$1.48\% & 44.41$\pm$0.93\% & 20.07$\pm$0.93\% & 30.87$\pm$0.56\% \\ 
GNN*  & Y & 32.33$\pm$0.52\% & 45.67$\pm$0.87\% & 22.50$\pm$0.67\% & 32.67$\pm$0.37\%  \\ 
    Closer Look*  & Y & 32.63$\pm$1.55\% & 43.76$\pm$0.93\% & 20.03$\pm$0.83\% & 30.67$\pm$0.36\% \\

\midrule

{PPN (Ours)}  & Y & 37.37$\pm$1.64\% & 50.31$\pm$1.00\% & 24.17$\pm$1.00\% & 36.21$\pm$0.57\%\\
{PPN+(Ours)}  & Y & \textbf{48.00$\pm$1.70}\% & \textbf{52.36$\pm$1.02}\% & \textbf{35.75$\pm$1.13}\% & \textbf{38.18$\pm$0.63}\%\\

\bottomrule

\end{tabular}
\label{table:exp_ws_pure}
\end{table*}

\section{Experiments}



\begin{table*}[th!]
\centering
\setlength{\tabcolsep}{3pt}
\caption{Validation accuracy (mean$\pm$CI$\%95$) on $600$ test tasks of PPN/PPN+ and baselines on WS-ImageNet-Mix.}
\begin{tabular}{lccccccc}
\toprule
\textbf{Model}   & \textbf{Weakly-Supervised}  &  \textbf{5way1shot} & \textbf{5way5shot} & \textbf{10way1shot} &  \textbf{10way5shot}  \\ \hline
Prototypical Net~\cite{snell2017prototypical} & N & 31.93$\pm$1.62\% & 49.80$\pm$0.90\% & 21.02$\pm$0.97\% & 36.42$\pm$0.62\%\\
GNN~\cite{gnn2018few} & N &  33.60$\pm$0.11\% & 45.87$\pm$0.12\% & 22.00$\pm$0.89\% & 34.33$\pm$0.75\%\\
Closer Look~\cite{closerlook}  & N & 33.10$\pm$1.57\% & 40.67$\pm$0.73\% & 20.85$\pm$0.92\% & 35.19$\pm$0.43\% \\

\midrule
Protytypical Net*  & Y & 31.80$\pm$1.48\% & 49.03$\pm$0.93\% & 20.33$\pm$0.98\% &34.79$\pm$0.58\% \\ 
GNN*  & Y & 30.33$\pm$0.80\% &  47.33$\pm$0.28\% & 23.33$\pm$1.03\% & 31.33$\pm$0.80\% \\ 
Closer Look*  & Y & 31.13$\pm$1.51\% & 44.90$\pm$0.78\% & 20.25$\pm$0.87\% & 34.01$\pm$0.40\% \\
\midrule
{PPN (Ours)}  & Y & 36.23$\pm$1.69\% & 52.38$\pm$0.92\% & 23.30$\pm$1.06\% & 38.20$\pm$0.55\%\\
{PPN+(Ours)}  & Y & \textbf{41.60$\pm$1.67}\% & \textbf{53.95$\pm$0.96}\% & \textbf{29.87$\pm$1.08}\% & \textbf{39.17$\pm$0.58}\%\\
\bottomrule

\end{tabular}
\label{table:exp_ws_mix}
\end{table*}

We compare PPN/PPN+ to three baseline methods, i.e., Prototypical Networks, GNN and Closer Look~\cite{closerlook}, and their variants of using the same weakly-labeled data as PPN/PPN+. 
For their variants, we apply the same level-wise training on the same weakly-labeled data as in PPN/PPN+ to the original implementations, i.e., we replace the original training tasks with level-wise training tasks. We always tune the initial learning rate, the schedule of learning rate, and other hyperparameters of all the baselines and their variants on a validation set of tasks.
The results are reported in Table~\ref{table:exp_ws_pure} and Table~\ref{table:exp_ws_mix}, where the variants of baselines are marked by ``*'' following the baseline name.

In PPN/PPN+ and all the baseline methods (as well as their variants), we use the same backbone CNN (i.e., $f(\cdot;\theta^{cnn})$) that has been used in most previous few-shot learning works~\cite{snell2017prototypical,finn2017model,vinyals2016matching}. It has 4 convolutional layers, each with $64$ filters of 3 $\times$ 3 convolution, followed by batch normalization~\cite{ioffe2015batch}, ReLU nonlinearity, and 2 $\times$ 2 max-pooling. 
The transformation $g(\cdot)$ and $h(\cdot)$ in the attention module are fully connected linear layers.

In PPN/PPN+, the variance of $\mP^0_y$ increases when the number of samples (i.e., the ``shot'') per class reduces. Hence, we set 
$\lambda=0$ in Eq.~(\ref{equ:p-final}) for $N$-way $1$-shot classifications, and $\lambda=0.5$ for $N$-way $5$-shot classification. 
During training, we lazily update $\mP^0_y$ for all the classes on the graph $\gG$ every $m=5$ epochs and choose the nearest $K=3$ neighbours as parents among all prototypes gained after training for PPN.
Adam~\cite{kingma2015adam} is used to train the model for 150k iterations, with an initial learning rate of $10^{-3}$, a weight decay of $10^{-4}$, and a momentum of $0.9$.
We reduce the learning rate by a factor of $0.7\times$ every 15k iterations starting from the 10k-th iterations.

\subsection{WS-ImageNet-Pure}
\label{sec:pure-exp}
WS-ImageNet-Pure is a subset of ILSVRC-12. 
On the ImageNet WordNet Hierarchy, we extract $~80\%$ classes from level-$7$ as leaf nodes of the category graph $\gG$ and use them as the targeted classes $\gY$ in few-shot tasks. The ancestor nodes of these classes on $\gG$ are then sampled from level-$6$ to level-$3$, which compose weakly-labeled classes $\gY^w$. 
We sub-sample the data points for classes on $\gG$ in a bottom-up manner: for any level-$7$ (bottom) level class $y$, we directly sample a subset from all the images belonging to $y$ in ImageNet; for any class $y$ on lower level-$j$ with $j<7$, we sample from all the images that belong to $y$ and have not been sampled into its descendant classes. Hence, we know that any data point sampled into class $y$ belongs to all the ancestor classes of $y$ but we do not know its labels on any targeted class of $y$. 
In addition, we sample each candidate data point for any class on level $j$ with probability $0.6^j$. 
Hence, the number of data points associated with each class thus reduces exponentially when the level of the class increases.
This is consistent with many practical scenarios, i.e., samples with finer-class labels can provide more information about targeted few-shot tasks, but they are much more expensive to obtain and usually insufficient. 

For training-test splitting of few-shot classes\footnote{Each training task is classification defined on a randomly sampled subset of training classes, while each test task is classification defined on a randomly sampled subset of test classes.}, we divide the classes from level-$7$ into two disjoint subsets with ratio $4$:$1$ ($4$ for training and $1$ for test). This ensures that any class in any test task has never been learned in training tasks. However, we allow training classes and test classes to share some parent classes. The detailed statistics of WS-ImageNet-Pure are given in Table~\ref{table:exp_ws_pure}.
\begin{table}[t!]
\caption{The average per-iteration time (in seconds) comparisons on PPN/PPN+ and other baselines on WS-ImageNet-Pure.}
\label{tab:time-cost}
\centering
\setlength{\tabcolsep}{3pt}
\begin{tabular}{|c|c|c|c|c|c|}
\hline
         & PPN & PPN+ & Proto Net & GNN & Closer Look\\ \hline
Training &  0.074 & 0.074 & 0.066 & 0.170 & 0.006 \\ \hline
Test     &  0.025 & 0.029 & 0.023 & 0.011 & 0.115  \\ \hline
\end{tabular}
\end{table}

The experimental results of PPN/PPN+ and all the baselines (and their weakly-supervised variants ending with ``*'') on WS-ImageNet-Pure are shown in Table~\ref{table:exp_ws_pure}. 
PPN/PPN+ outperform all other methods. The table shows that PPN/PPN+ are more advantageous in 1-shot tasks, and that PPN+ achieves $\sim15\%$ improvement compared to other methods. This implies that the weakly-supervised information can be more helpful when supervised data is highly insufficient, and our method is able to significantly boost performance by exploring the weakly-labeled data. Although all the weakly-supervised variants of baselines are trained on the same data as PPN/PPN+, they do not achieve similar improvement because their model structures do not have such mechanisms as the prototype propagation in PPN which relates different classes and tasks. In addition, training on unrelated tasks can be distracting and even detrimental to performance. In contrast, PPN/PPN+ build a computational graph of prototypes associated with both coarse and fine classes, and the error on any class can be used to update the prototypes of other classes via backpropagation on the computational graph. 


\subsection{WS-ImageNet-Mix}
\label{sec:mix-exp}
To verify if PPN can still learn from weakly-labeled data that belong to other fine classes not involved in the few-shot tasks, we propose another subset of ILSVRC-12, WS-ImageNet-Mix, whose detailed statistics are given in Table~\ref{tab:datasets}. 
We extract WS-ImageNet-Mix by following the same procedure as extracting WS-ImageNet-Pure except that: 1) data points sampled for a coarse (non-leaf) class can belong to the remaining $\sim20\%$ level-$7$ classes outside of the $\sim80\%$ level-$7$ classes used for generating few-shot tasks; and 2) for any class on level-$j$, we sample each data point with probability $0.7^j$ instead of $0.6^j$.



The experimental results are reported in Table~\ref{table:exp_ws_mix}, which shows that PPN/PPN+ still outperform all baselines, and PPN+ outperforms them by $\sim10\%$ for 1-shot classification. This indicates that PPN/PPN+ is robust to (and might be able to leverage) weakly-labeled data unrelated to the final few-shot tasks.


\subsection{Effects of Prototype Propagation}
\begin{figure}[t!]
\begin{center}
\includegraphics[width=\linewidth]{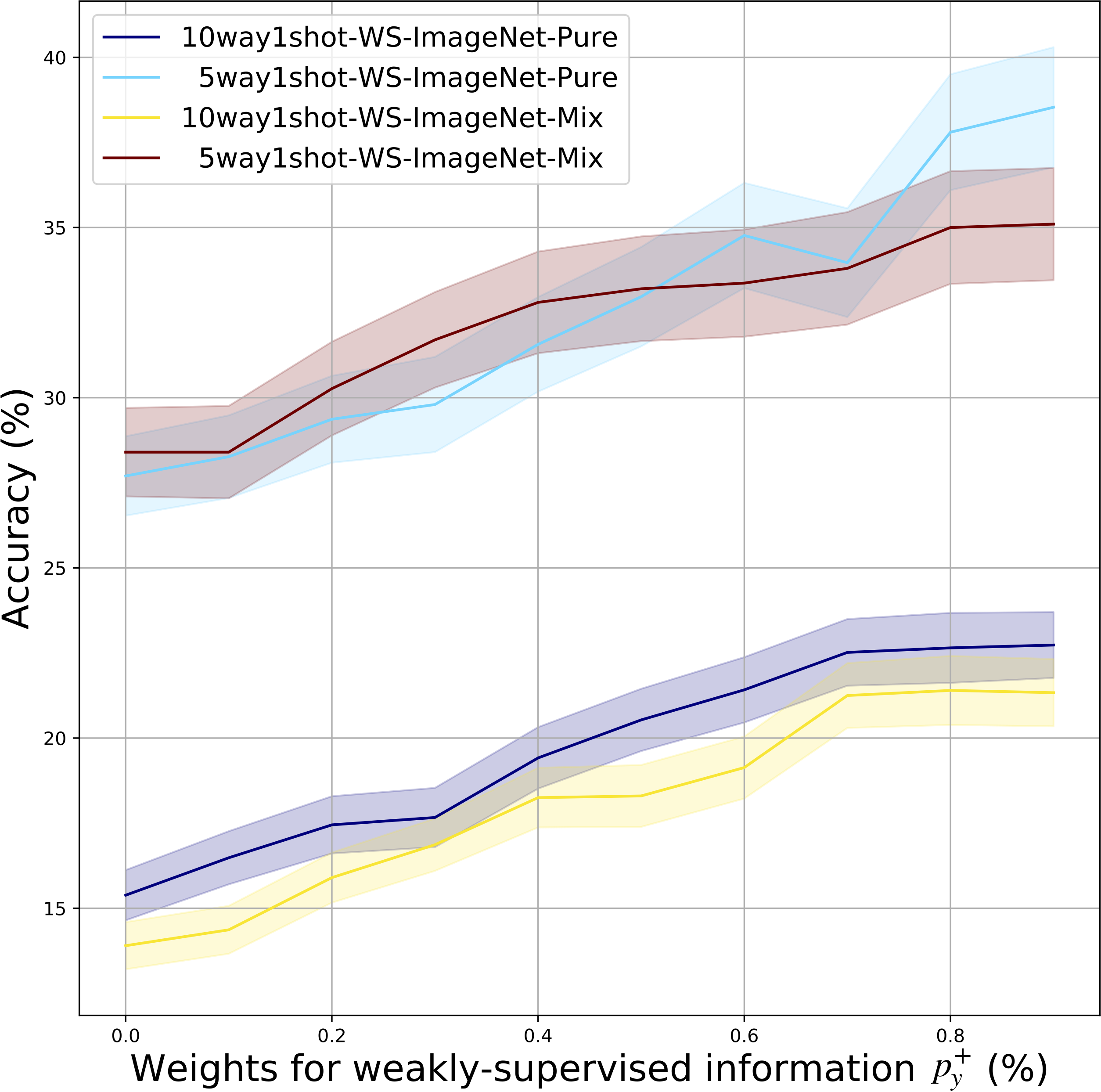}
\end{center}
\caption{Validation accuracy averaged over $600$ $N$way1shot test tasks improves when the weight of $\mP_{y}^{+}$ i.e., $1-\lambda$ in Eq.~(\ref{equ:p-final}) increases, which implies the effectiveness of prototype propagation.}
\label{fig:ratio-acc}
\end{figure}
To study whether and how the propagation in Eq.~(\ref{equ:p-par-hop1}) improves few-shot learning, we evaluate PPN using different $\lambda$ in Eq.~(\ref{equ:p-final}). 
Specifically, we try different weights $1-\lambda$ (x-axis in Figure~\ref{fig:ratio-acc}) for $\mP_{y}^{+}$ in Eq.~(\ref{equ:p-final}) between $[0,0.9]$, and report the validation accuracy (y-axis in Figure~\ref{fig:ratio-acc}) on test tasks for $N$way1shot tasks and the two datasets. 
In all scenarios, increasing the weight of $\mP_{y}^{+}$ in the given range consistently improves the accuracy (although the accuracy might drop if the weight gets too close to $1$ though), which demonstrates the effectiveness of prototype propagation.

\subsection{Time Cost}

The average per-iteration time (in seconds) on a single TITAN XP for PPN/PPN+ and the baselines are listed as Table~\ref{tab:time-cost}. PPN has moderate time cost comparing to other baselines. Compared to prototypical network, our propagation procedure only requires $\sim$10\% extra time cost but significantly improves the performance.

\section{Future Studies}
In this work, we propose to explore weakly-labeled data by developing a propagation mechanism generating class prototypes to improve the performance of few-shot learning. Empirical studies verifies the advantage of exploring the weakly-labeled data and the effectiveness of the propagation mechanism.
Our model can be extended to multi-steps propagation and assimilate more weakly-labeled information. 
The graph can be more general than a class hierarchy, and the node on the graph is not limited to be a class: it can be extended to any output attribute.\looseness-1

\section{Acknowledgements}
This research was funded by the Australian Government through the Australian Research Council (ARC) under grants 1) LP160100630 partnership with Australia Government Department of Health and 2) LP150100671 partnership with Australia Research Alliance for Children and Youth (ARACY) and Global Business College Australia (GBCA). We also acknowledge the support of NVIDIA Corporation and MakeMagic Australia
with the donation of GPUs.

\bibliographystyle{named}
\bibliography{ms}

\end{document}